\documentclass[runningheads]{llncs}

\usepackage{eccv}

\usepackage{eccvabbrv}

\usepackage{graphicx}
\usepackage{wrapfig}
\usepackage{booktabs}
\usepackage{siunitx}
\usepackage{multirow}
\usepackage[final]{microtype}
\usepackage[accsupp]{axessibility}  

\usepackage{subcaption}
\usepackage{booktabs}
\usepackage{multirow}
\usepackage{siunitx}

\usepackage{hyperref}

\usepackage{orcidlink}

\newcommand{\ignore}[1]{}

\begin{document}

\title{Explainability-Aware Frustum Attack: Exposing Structural Vulnerabilities in LiDAR-Based 3D Object Detectors}

\titlerunning{Saliency-LiDAR}

\author{Chengzeng You\inst{}\orcidlink{0000-0002-9558-4584} \and
Binbin Xu\inst{}\orcidlink{0000-0002-1967-3219}\thanks{Now at Google.} \and
Soteris Demetriou\inst{}\orcidlink{0000-0003-0318-9171}}

\authorrunning{C.~You et al.}

\institute{Imperial College London, UK\\
\email{\{chengzeng.you19, b.xu17, s.demetriou\}@imperial.ac.uk}
\\
}

\maketitle

\begin{abstract}

The structural vulnerabilities of point cloud–based 3D object detectors remain poorly understood. Prior work has studied adversarial robustness primarily on isolated 3D object models, while recent LiDAR spoofing attacks target richer and more realistic driving scenes but focus mainly on physical realizability rather than understanding detector behavior or attack efficiency. In this work, we investigate how LiDAR-based detectors rely on spatial evidence in complex scenes and whether these reliance patterns can be exploited to induce failures more efficiently.
To this end, we propose an explainability-guided adversarial analysis methodology. We introduce the Saliency-LiDAR (\textit{SALL}) method, which aggregates Integrated Gradient attributions across scenes to produce universal saliency maps for LiDAR-based 3D object detectors. Guided by these maps, we design the Explainability-aware Frustum Attack (\textit{EFA}), which selectively perturbs only the most influential frustums rather than uniformly attacking entire object regions. Experiments on KITTI and nuScenes, across detectors such as PointPillars and SECOND, show that EFA reduces detection recall by more than 15 percentage points while requiring 25–50\% fewer perturbed frustums than the SOTA non–saliency-aware baseline. These findings reveal that modern 3D detectors concentrate discriminative evidence in a small subset of spatial regions, exposing a structural robustness vulnerability in current LiDAR perception systems. Our code is released at 
\url{https://github.com/SecMindLab/Saliency_LiDAR}.


\end{abstract}

%

\section{Introduction}
Understanding the vulnerabilities of modern 3D object detectors is 
increasingly important as these models are deployed in real-world 
perception systems. While significant progress has been made in 
accuracy, far less is known about the structural dependencies that 
govern their decisions and how attribution patterns affect 
robustness. Prior explainability-driven work has begun probing this question. 
Tan \etal~\cite{tan2023explainability} showed that, for isolated 
3D object classifiers operating on normalized CAD models \cite{wu20153d}, a sparse 
set of critical points can strongly influence predictions. However, 
such studies are limited to digital classification settings and 
do not address full-scene 3D object detection pipelines, where 
localization, proposal generation, non-maximum suppression, clutter, 
occlusion, and range-dependent sparsity fundamentally alter the 
decision structure.

Autonomous driving presents both a more critical and more challenging 
testbed. LiDAR-based 3D detectors are central to safety-critical 
systems increasingly deployed in urban environments. At the same time, 
driving scenes contain multiple interacting objects, background clutter, 
sensor sparsity, and geometric constraints absent from simplified 
benchmarks \cite{wu20153d}. Understanding structural failure modes in this setting is 
therefore both timely and methodologically more challenging 
than simplified digital benchmarks. Prior work in this domain has primarily studied LiDAR spoofing and 
frustum-level attacks \cite{cao2023you,sato2024lidar,sato2025realism}. 
These methods demonstrate that manipulating or removing entire frustums 
can hide objects, but typically require perturbing large spatial regions. 
They do not explicitly analyze whether a detector's reliance is 
concentrated in smaller, structurally critical regions, nor whether such 
regions are consistent across scenes and models.

\begin{figure}[tb]
\centering
    \begin{subfigure}{0.45\linewidth}
    \centering
        \includegraphics[clip, trim=0.0cm 3cm 0.0cm 0.0cm,width=0.95\linewidth]{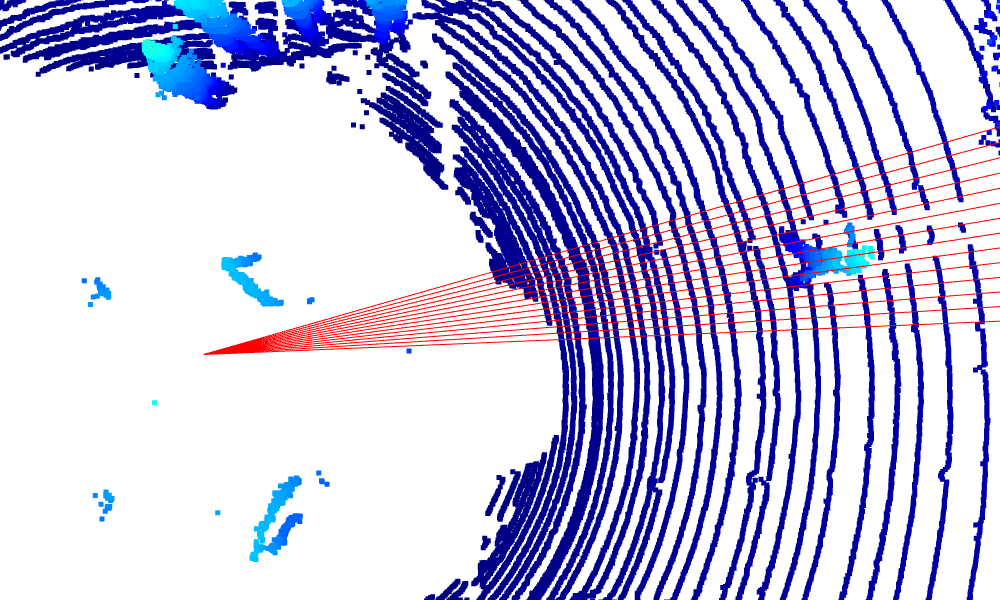}
        \caption{Previous attacks \cite{cao2023you,sato2024lidar,sato2025realism} on pedestrian.}
        \label{fig:ex_lidar_ped_previous}
    \end{subfigure}
    \begin{subfigure}{0.45\linewidth}
    \centering
        \includegraphics[clip, trim=0.0cm 3cm 0.0cm 0.0cm,width=0.95\linewidth]{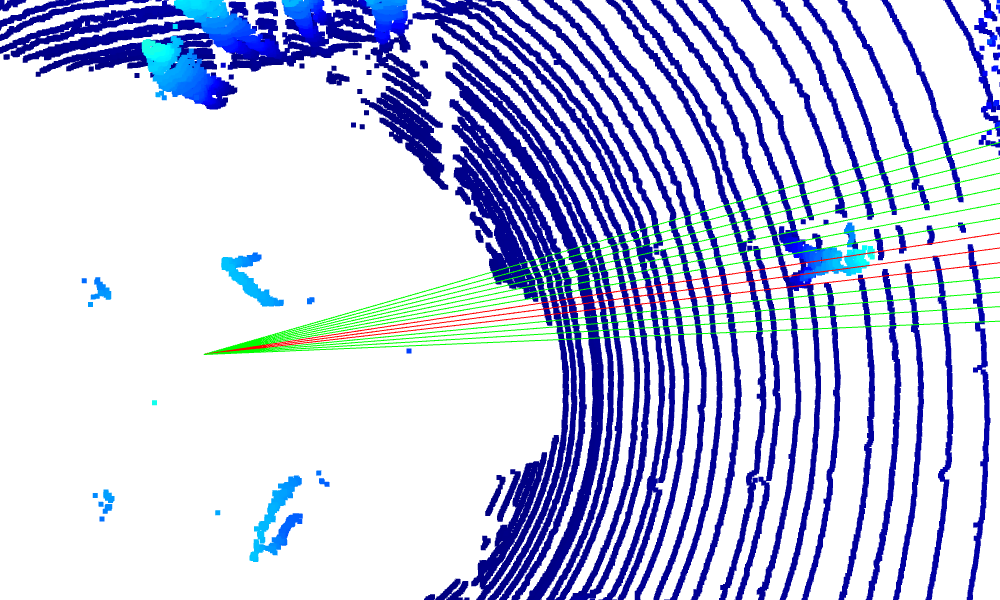}
        \caption{Our attack on pedestrian.}
        \label{fig:ex_lidar_ped_ours}
    \end{subfigure}
    \caption{Saliency-guided hiding visualization. Red rays denote perturbed 
    frustums while green rays denote unperturbed frustums. Prior approaches remove all frustums associated with the object, whereas our method perturbs only a few critical frustums.}
    \label{fig: saliency lidar visualization}
\end{figure}

In this work, we bridge explainability and LiDAR attack research by 
introducing \textbf{SALL}, a framework that aggregates instance-level 
Integrated Gradients into \emph{universal, class-level} saliency maps for 
LiDAR-based 3D object detection. These maps reveal stable spatial importance distributions across rich real-world 3D scenes, from different datasets, and when derived from distinct detector architectures. 
We then use these saliency priors to construct a structured perturbation 
strategy, \textit{EFA}, operating under geometry-consistent constraints 
inspired by LiDAR sensing.
\cref{fig: saliency lidar visualization} illustrates the key 
distinction. Whereas prior attacks 
\cite{cao2023you,sato2024lidar,sato2025realism} perturb all frustums 
associated with an object, our method enables targeting only the most salient 
regions as identified by \textit{SALL}. Our evaluation shows that 
universal saliency maps are 
transferable across datasets collected in real-world driving scenarios and key detector architectures. Under 
geometry-consistent perturbation constraints, targeting saliency-identified regions, our \textit{EFA} achieves over \SI{95}{\percent} degradation rates, 
improving effectiveness by \SIrange{20}{30}{\percent} on large objects 
and approximately \SI{23}{\percent} on small objects compared to prior 
frustum-based approaches 
\cite{cao2023you,sato2024lidar,sato2025realism}, while affecting 
\SIrange{25}{50}{\percent} fewer regions. 

\noindent{}In summary, our main contributions are as follows:
\begin{itemize}
    \item We propose \textbf{SALL}, a framework that aggregates 
    instance-level attributions into universal, class-level saliency 
    maps for LiDAR-based 3D object detection.
    \item We demonstrate that these maps reveal consistent discriminative regions which transfer across detector architectures and datasets 
    in realistic autonomous driving scenarios.
    \item We introduce a geometry-consistent saliency-guided frustum selection and perturbation 
    strategy (\textit{EFA}) and show that it substantially outperforms 
    prior frustum-based methods~\cite{cao2023you,sato2024lidar,sato2025realism} while requiring 
    significantly smaller perturbation budgets.
\end{itemize}

\section{Related Work}
\label{sec:related}

\subsection{LiDAR Spoofing Attacks}
\label{sec: lidar spoofing attacks}
LiDAR measurements can be spoofed by replaying LiDAR pulses to create fake points in the sensed environment. Such an attack is challenging for the LiDAR system to recognize as it doesn't require any physical contact with the LiDAR sensor or interference with the processing of sensor measurements. To perform realistic attacks, researchers have been improving the hardware and software of LiDAR spoofers\cite{shin2017illusion, sun2020towards, sato2024lidar, jin2023pla, sato2025realism, cao2023you}. A common attack strategy is to capture LiDAR signals from the victim LiDAR, then add a time delay and fire fake laser beams back to the victim LiDAR. Fake points are shown to be reliably injected to fool 3D object detectors to output erroneous predictions. As a result, real objects can be hidden \cite{tu2020physically, modas2020toward,hau2021object,jin2023pla,you2024adversarial,sato2024lidar,cao2023you, sato2025realism} while ghost objects can be injected \cite{sun2020towards,hallyburton2022security,jin2023pla, sato2024lidar}. Real object hiding is regarded as a more dangerous type of attack than ghost object injection, as it is more likely to cause fatal collisions. Based on digital effects, object hiding attacks can be categorized into 2 types of attacks: point-level hiding attacks and frustum-level hiding attacks. For point-level hiding attacks, the adversary 
leveraged spoofers, attack algorithms or 3D-printed objects \cite{tu2020physically,modas2020toward} to shift \cite{hau2021object,you2024adversarial} or introduce a few points \cite{jin2023pla} in the view of LiDAR-based perception.  Frustum-level hiding attacks \cite{cao2023you,sato2024lidar, sato2025realism}, by taking advantage of recent spoofing hardware improvements, have been shown to be more powerful than point-level hiding attacks.  Specifically,  the adversary is equipped with the capability to perturb the whole frustum space to successfully hide objects. Once the frustum is targeted, all points inside the frustum are perturbed. 

This creates a clear trade-off: frustum-level attacks are powerful but highly inefficient as they require a large spoofing area that can lead to hardware overheating~\cite{sato2025realism}; while point-level attacks are efficient but less powerful. Our work bridges this gap by proposing a saliency-guided frustum selection and perturbation strategy to achieve the effectiveness of a full frustum attack with the operational efficiency of a point-level one.


\subsection{Critical Points and Saliency Map}

Identifying \textit{which} regions to attack is a central challenge in our work. Prior research has shown that 
critical points\cite{qi2017pointnet} contribute to features of max-pooling layers and summarize skeleton shapes of input objects \cite{tan2023explainability}. Based on critical points, researchers further studied the model robustness by perturbing or dropping critical point set identified through monitoring the max-pooling layer or accumulating losses of gradients \cite{kim2021minimal, yang2019adversarial, zheng2019pointcloud}. However, capturing the output of the max-pooling layer struggled to identify discrepancies between key points, and simultaneously, saliency maps based on raw gradients have been proven to be defective \cite{adebayo2018sanity,sundararajan2016gradients}. To overcome these issues, Tan \etal\cite{tan2023explainability} introduced Integrated Gradient (IG) \cite{sundararajan2017axiomatic} which is oriented on generating saliency maps of inputs by calculating gradients during propagation, to investigate the sensitivity of model robustness to the critical point sets and successfully fooling target classifiers with very few point perturbations. However, this study identified critical points specific to object point clouds \cite{wu20153d} without further summarizing richer 3D scenes \cite{geiger2013vision, caesar2020nuScenes}. As a result, for every instance of an object, the attacker needs to run a separate iterative optimization process. 
Zhu \etal\cite{zhu2021can} proposed an attack framework based on which the attacker could identify a few adversarial locations in the 3D LiDAR scene. They further studied the characteristics of adversarial locations by visualizing them using region maps. However,  the region map generation using digital point injection attacks failed to deal with point occlusion problems which disobeyed the physics of LiDAR. This results in reduced reliability of the region map. Additionally, the region map is not considered representative due to limited samples (100 samples), one single dataset and one type of object (car object). Our work, \textit{SALL}, builds on these concepts but differs in three fundamental ways:
\begin{enumerate}
    \item \textbf{Task:} 
    Inspired by the  high-level idea of gradient attribution in the task of  single-object classification, we design \textit{SALL} for the more complex domain of \emph{3D object detection} in large-scale autonomous driving scenes.
    \item \textbf{Universality:} Unlike methods requiring per-instance optimization~\cite{tan2023explainability}, \textit{SALL} aggregates saliency maps across thousands of complex, real-world LiDAR scenes  and objects to derive a \emph{universal saliency map} for an entire object class, despite immense instance variability (direction, size, occlusion, \etc).
    \item \textbf{Realism \& Robustness:} Unlike region maps built on unrealistic physics~\cite{zhu2021can}, our maps are generated using physically-aware perturbations and are validated across multiple large-scale datasets 
    and object types.
\end{enumerate}

\section{Methodology}
\subsection{Threat Model}
\label{Threat Model}
We consider a \emph{geometry-constrained adversary} ($\mathcal{A}$) that 
modifies LiDAR point clouds under constraints consistent with LiDAR sensing. 
The adversary’s capabilities are defined to align with behaviors demonstrated in prior LiDAR spoofing and signal manipulation works  
\cite{petit2015remote,shin2017illusion,cao2019adversarial,sun2020towards,
sato2024lidar,cao2023you,sato2025realism,hau2021object}. Specifically, $\mathcal{A}$ may remove, shift, or displace a subset of points 
along their original LiDAR ray directions. Such perturbations correspond to 
physically plausible manipulations of return timing, where points may appear 
either nearer~\cite{shin2017illusion} or further 
\cite{cao2019adversarial,hau2021object} relative to the sensor. We assume 
these operations can be applied to multiple rays and maintained over time 
as the platform moves~\cite{cao2019adversarial,sun2020towards,sato2024lidar}.


$\mathcal{A}$ is restricted to localized spatial regions (e.g., angular frustums) under a limited perturbation budget and cannot arbitrarily rewrite the point cloud, introduce globally inconsistent structures, or manipulate other sensing modalities. These constraints preserve ray consistency and local geometric structure while enabling structured adversarial evaluation.

Following Hau \etal~\cite{hau2021object}, we assume $\mathcal{A}$ has 
query access to victim detector outputs (\eg, confidence scores) but no access to model architecture or parameters. 
The attack therefore operates in a black-box setting with respect to model 
internals. We also evaluate a version of $\mathcal{A}$ without knowledge of which detector the victim had deployed. 
The objective of $\mathcal{A}$ is to reduce detection confidence and induce 
missed detections through geometry-consistent perturbations. Our design goal is 
not to emulate a specific spoofer, but to analyze 3D detector sensitivity 
under structured perturbations grounded in prior empirical demonstrations.

\subsection{Saliency-LiDAR and Saliency Maps}
\label{sec:sall}
\begin{figure}[tb]
    \centerline{\includegraphics[clip, trim=0.0cm 0cm 0.0cm 0cm, width=\linewidth]{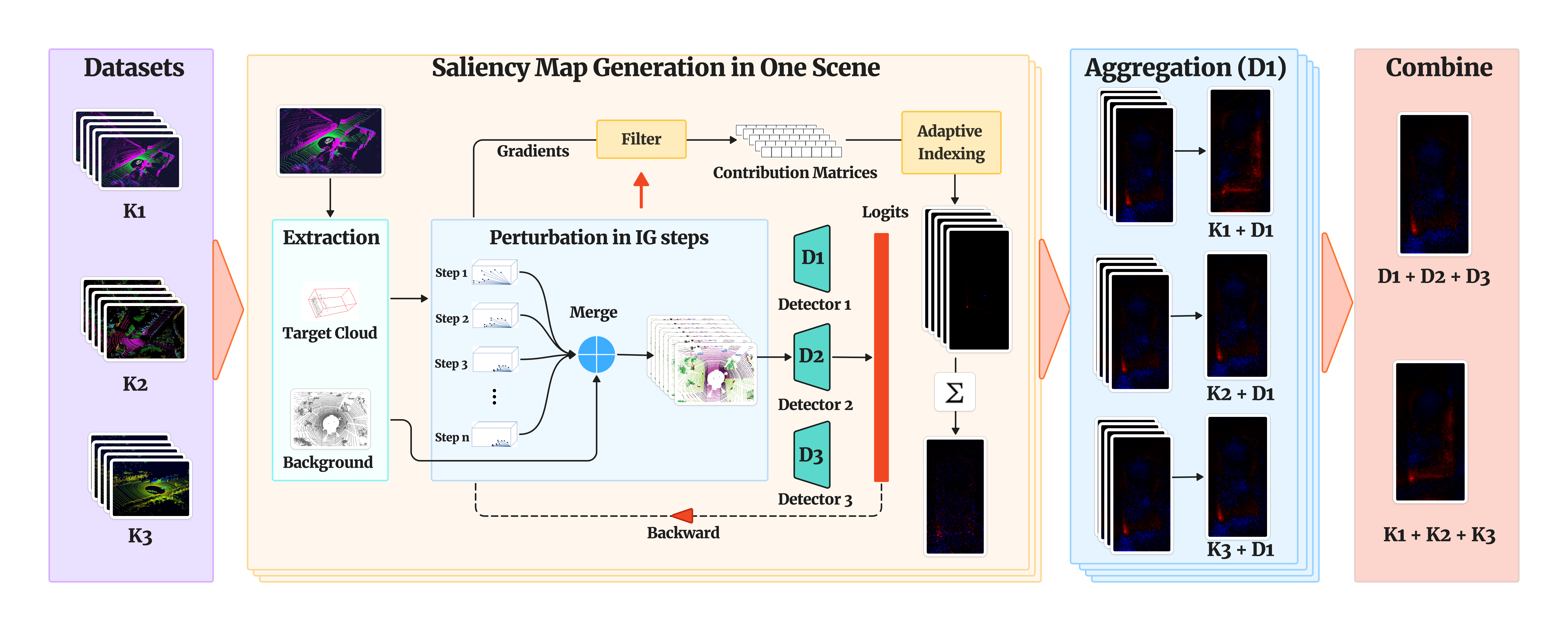}}
    \caption{Overview of universal saliency map generation for LiDAR objects with \textit{SALL}. In \textit{Filter} operation, only gradients of the best predictions are saved by referring to the perturbation region. In \textit{Adaptive Indexing}, point-level saliency maps are downsampled to 2D matrices. $\Sigma$ denotes the summation of 2D matrices. }
    \label{fig:saliency_lidar}
    \vspace{-15pt}
\end{figure}

To identify critical regions for 3D object detection, we propose Saliency-LiDAR (SALL). Inspired by Tan et al~\cite{tan2023explainability}, SALL adapts Integrated Gradients to autonomous-driving object detection and aggregates instance-level attributions across scenes into universal saliency maps for each object class. ~\cref{fig:saliency_lidar} summarizes the pipeline.

\noindent\textbf{Preprocessing.} \textit{SALL} takes a raw 3D scene $S$ as input. Before IG computation, it preprocesses the scene through an extraction module ($\mathcal{E}$). After extracting target object clouds $T$ and background points $G$, the target object clouds $T$ are fed into the IG component to compute point-level contributions. 

\noindent\textbf{Integrated Gradient Computation.}  For each IG step, points in a $T^i$ are perturbed by a perturbation module ($\mathcal{P}$)  and outputs ${T^i}'$.  Given a base point $b_0 = (x_0, y_0)$, each point $p_j = (x_j, y_j)$ in $T^i$ is first converted to a vector $\vec{z_j} = (x_j - x_0, y_j - j_0)$. The base point is customized. For example, $\mathcal{A}$ can select the LiDAR unit as the base point to obey the physics of LiDAR. $\mathcal{A}$ can also select the corner point of the focus region $R$  to only perturb points inside the region and make better use of each IG step.

Let the overall IG steps be $M$, at the N-th IG step ($N \leq M$), perturbed point ${p_j}'$ in ${T^i}'$ can be calculated in \cref{eq:4}.

\begin{equation}
    \label{eq:4}
    {p_j}' = b_0 + \frac{N}{M}*\vec{z_j}
\end{equation}

All ${T^i}'$ in perturbed object clouds $T'$ are then merged with the background points ($G$) to produce the perturbed 3D scene ($S'$): $S' = T' \bigoplus G$. 
$S'$ is used for the gradient computation. It passes through a 3D object detector ($\mathcal{D}$) which outputs a set of logits $O^i$ for each target object $i$. To focus on target objects instead of the whole LiDAR scene,  Intersection of Unions (IoUs) between the focus regions $R$ and predicted bounding boxes are first computed to identify the best predictions.  Gradients of the best predictions are saved while other gradients are filtered out in the filter module. After that, a point-level contribution map $C^p_i$ is generated per target object. 
Finally, an integrator function integrates all $C^p_i$ across all IG steps,
to produce a point-level saliency or contribution map $C^p$ for objects in a single scene. 

\noindent\textbf{Adaptive Indexing.} 
Since $R$ regions have different dimensions and rotations in different LiDAR scenes, to generate a universal saliency map, point-level saliency maps need to be downsampled to pixel-level with the same size. To achieve that,  each extracted target point cloud $T^i$ is first converted from LiDAR coordinates to bounding box coordinates. Then, given the target size of the universal saliency map (2D-pixel image),  each target object's voxel size is adaptively computed based on $R$'s dimension. After that, indices of each point in $T^i$ can be computed. According to the point-level saliency map $C^p$, the contribution of each voxel is summed up to generate a 2D-pixel matrix $C^v$ in which each element indicates the contribution of each voxel. 

\noindent\textbf{Aggregation Across Scenes.} For each scene $S$, we generate a contribution matrix $C^v$. $C^v$s are then aggregated across all $k$ scenes by simple matrix additions to generate the universal saliency map $C^{uv}$ for the target object type. 

\noindent\textbf{Combining Different Detectors and Datasets.} On the benefit of \textit{SALL}'s aggregation nature, $C^{uv}$s learned from different datasets (\eg, KITTI \cite{geiger2013vision} or nuScenes \cite{caesar2020nuScenes}) and different detectors (\eg, PointPillars \cite{lang2019pointpillars} or SECOND \cite{yan2018second}) can also be combined to generate a joint universal saliency map which is more representative than the individual saliency map. 

\begin{figure}[tb]
\captionsetup[subfigure]{font=scriptsize}
\centering
    \begin{subfigure}{0.4\linewidth}
    \includegraphics[width=\linewidth]{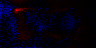}
        \caption{PointPillars in KITTI.}
        \label{fig:kitti_pp_car}
    \end{subfigure}
    \begin{subfigure}{0.4\linewidth}
        \includegraphics[width=\linewidth]{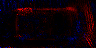}
        \caption{SECOND in KITTI.}
        \label{fig:kitti_second_car}
    \end{subfigure}
    
    \begin{subfigure}{0.4\linewidth}
        \includegraphics[width=\linewidth]{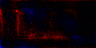}
        \caption{PointPillars  in nuScenes.}
        \label{fig:nusc_pp_car}
    \end{subfigure}
    \begin{subfigure}{0.4\linewidth}
        \includegraphics[width=\linewidth]{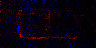}
        \caption{SECOND in nuScenes.}
        \label{fig:nusc_second_car}
    \end{subfigure}
    
    \begin{subfigure}{0.4\linewidth}
        \includegraphics[width=\linewidth]{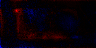}
        \caption{PointPillars  in KITTI  + nuScenes.}
        \label{fig:nusc_kitti_pp_car}
    \end{subfigure}
    \begin{subfigure}{0.4\linewidth}
        \includegraphics[width=\linewidth]{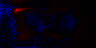}
        \caption{PointPillars  + SECOND in KITTI.}
        \label{fig:kitti_pp_second_car}
    \end{subfigure}
    \caption{ Saliency map visualization of \emph{Cars} in different datasets and detectors. Red pixels denote positive contribution values while blue pixels denote negative contributions.}
    \label{fig:saliency map visualization}
    \vspace{-10pt}

\end{figure}

\noindent\textbf{Saliency Map Visualization.}
\cref{fig:saliency map visualization} shows the saliency map for $Car$ objects at 5 - 8m.  Most positive pixels fall in edges with some less critical and negative pixels falling in the center of the bounding box. This is true for LiDAR objects where most points appear on the surfaces. Specifically, PointPillars-based saliency maps tend to highlight the left corner as the most critical area while SECOND-based saliency maps prefer to indicate a $L~Shape$ area that aligns with bounding box edges. It is also obvious that the joint saliency maps across different datasets (\cref{fig:nusc_kitti_pp_car}) and different detectors (\cref{fig:kitti_pp_second_car}) become more representative by strengthening common areas and weakening uncommon areas.

\subsection{Explainability-Aware Frustum Attack}

\label{sec:EFA}
Explainability-aware Frustum Attack (\textit{EFA}) consists of six phases, taking in raw LiDAR point cloud ($S$) of the scene, performing perturbation of target objects, and producing the adversarial point cloud ($S'$) as its output.

\noindent\textbf{Phase 1: Extraction.} \textit{EFA} detects objects from $S$.
Then it separates $S$ into background points $G$ ($G \in \mathbb{R}^{n_g \times d}$) and a set of target point clouds
$T=\{T^1, T^2,...,T^m\}$.
Each target point cloud $T^i$ is extracted from a region of interest $R$. 
$R$ can be defined by the ground truth bounding box to solely focus on the exact object cloud or by the expanded dimension bounding box to include surrounding points of the detected object.

\noindent\textbf{Phase 2: 2D Indexing.} Each target point cloud $T^i$
is further discretized. Inspired by Lang \etal~\cite{lang2019pointpillars}, 
we compute the corresponding indices of each point in pillar format.
2D indexing is more efficient than voxelisation methods,
because it does not need to convert points to voxels.
Also, the corresponding voxel size is customized and can be set to near point level where each voxel only contains a few points or even one point. Based on the indices, $\mathcal{A}$ can define a virtual patch as shown in \cite{you2024adversarial} or get critical values according to the coordinates of the saliency map generated by our \textit{SALL} method (\cref{sec:sall}).

\noindent\textbf{Phase 3: Frustum Simulation.} Given LiDAR's origin $P^0 = (x_0, y_0, z_0)$ , each point $ P_j = (x_j, y_j, z_j)$ in $T^i$ is converted into a 2D degree $D_j$ in \cref{eq:1}.
\begin{equation}
\label{eq:1}
    D_j = \arctan\frac{(y_j - y_0)}{(x_j - x_0)}*\frac{180^\circ}{\pi}
\end{equation}
$R$'s corners are also taken into account for comparison to find a minimum degree $D_{min}$. Assume the frustum degree is $D_{step}$, the relative frustum index $f_{id}$ can be calculated in \cref{eq:2}


\begin{equation}
    \label{eq:2}
    f_{id} = \left\lfloor \frac{D_j - D_{min}}{D_{step}} \right\rfloor
\end{equation}

\noindent\textbf{Phase 4: Frustum Selection.} Based on frustum indices, there are 2 frustum selection strategies: $Random~Frustum~Selection$ and $Critical~Frustum~First$.  With $Random~Frustum~Selection$, given a frustum-perturbation budget, $\mathcal{A}$ randomly selects frustums among frustum candidates. With $Critical~Frustum~First$, $\mathcal{A}$ selects the most critical frustums according to the \textit{SALL}-generated universal saliency map of the target object. The criticality of each frustum is determined by the criticality of all points inside the frustum. Because of \textit{EFA}'s modular architecture, other novel selection strategies can also be easily incorporated.

\noindent\textbf{Phase 5: Frustum Perturbation.} After selecting frustums, different perturbation strategies can be applied to perturb points inside each frustum. For example, \textit{EFA} supports the $Remove~Perturbation$ strategy similarly to PRA \cite{cao2023you} which simply removes all points within a frustum. \textit{EFA} also supports $Shift~Perturbation$ strategy in \cite{you2024adversarial} which shifts points in accordance with the rays that the LiDAR points fall on.  The result is a perturbed point cloud $V'$. 

\noindent\textbf{Phase 6: Merge.} All $V'$s are then merged with 
$G$ to output the final 
adversarial 3D LiDAR scene $S' = G \bigoplus V'$.
$S'$ is in the same format as the original LiDAR scene $S$, 
and can be fed into any LiDAR-based detectors for evaluations. 

\begin{figure}[tb]
\centering
    \begin{subfigure}{0.24\textwidth}
        \includegraphics[width=\linewidth]{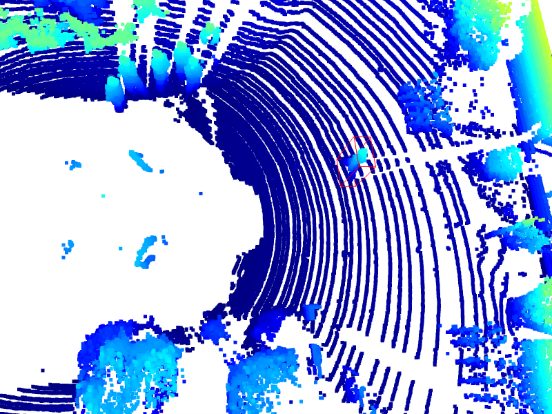}
        \caption{Benign scene.}
        \label{fig:benign lidar scene}
    \end{subfigure}
    \hfill
    \begin{subfigure}{0.24\textwidth}
        \includegraphics[width=\linewidth]{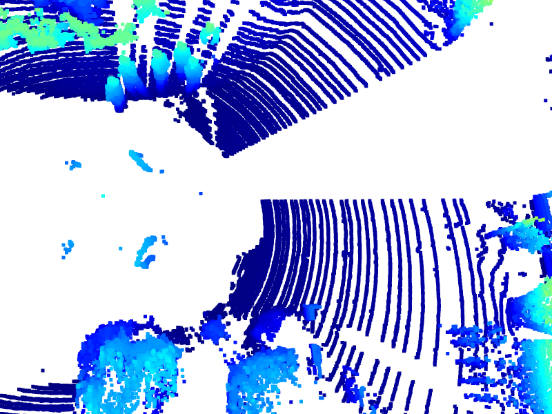}
        \caption{PRA \cite{cao2023you}}
        \label{fig: pra}
    \end{subfigure}
    \hfill
    \begin{subfigure}{0.24\textwidth}
        \includegraphics[width=\linewidth]{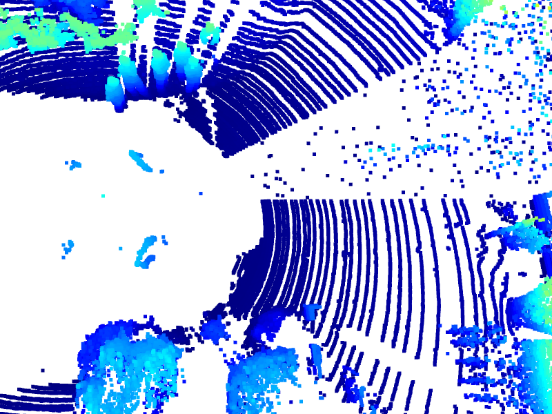}
        \caption{HFR \cite{sato2024lidar}}
        \label{fig: hfr}
    \end{subfigure}
    \hfill
    \begin{subfigure}{0.24\textwidth}
        \includegraphics[width=\linewidth]{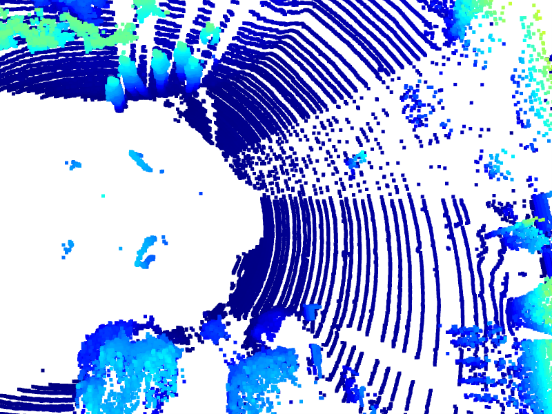}
        \caption{A-HFR \cite{sato2025realism}}
        \label{fig: a-hfr}
    \end{subfigure}
    \caption{Instantiation of prior frustum-level LiDAR attacks. Our simulations of attacking a 30\textdegree{} frustum area can achieve the same digital attack effects as \cite{cao2023you, sato2024lidar, sato2025realism}.}
    \label{fig:baseline_realization}
    \vspace{-10pt}
\end{figure}

\noindent\textbf{Instantiation of Prior Frustum-Level Perturbations.} To enable controlled comparisons, we digitally instantiate the 
perturbation strategies proposed in prior frustum-level LiDAR attacks, including PRA~\cite{cao2023you}, HFR~\cite{sato2024lidar}, and A-HFR~\cite{sato2025realism}. 
For alignment with prior works, perturbations are applied to the central  region of the target object. HFR is instantiated using 
\textit{Shift Perturbation}, PRA using \textit{Remove Perturbation}, and A-HFR using \textit{Remove Perturbation} with point removal rates reported in~\cite{sato2025realism}. All methods operate over a 
30\textdegree{} frustum span to ensure comparable perturbation budgets. As illustrated in \cref{fig:baseline_realization}, these 
instantiations reproduce the characteristic digital effects reported in~\cite{cao2023you,sato2024lidar,sato2025realism}. In subsequent experiments, we use these digitally instantiated baselines to compare against our saliency-guided strategy (\textit{EFA}) under identical frustum 
constraints.

\section{EFA Evaluation}
\subsection{Experimental Setup}
\label{sec:setup}

\textbf{Datasets.} We use the KITTI~\cite{geiger2013vision} trainval dataset (\num{7481} scenes) and the nuScenes~\cite{caesar2020nuScenes} trainval dataset (\num{34149} samples) for saliency map generation and evaluation.

\noindent\textbf{Target Objects.} We focus on $Car$ objects within \SIrange{5}{8}{\meter} in front of the ego-vehicle. Due to the lower frequency of $Pedestrian$ objects, we expanded their focus region to \SIrange{0}{20}{\meter}. For each object, we define the Region of Interest ($R$) by expanding its ground-truth bounding box by a scale factor of 1.5 (as described in \cref{sec:EFA}) to understand the contribution of surrounding points.

\noindent\textbf{Target Models.} We evaluate our attack against two popular and representative 3D object detectors: PointPillars~\cite{lang2019pointpillars} and SECOND~\cite{yan2018second}.

\noindent\textbf{Attack Parameters.} We define our \textit{EFA} using a frustum degree $D_{step} = \ang{1}$, matching the average resolution of LiDARs attacked in prior work~\cite{sato2024lidar}. For \textit{SALL}, we set the number of Integrated Gradient steps to $M=25$ and use the nearest corner of $R$ as the IG base point (see \cref{sec:sall}). Our adaptive indexing module normalizes all saliency maps to a fixed size of $64 \times 32$ voxels, resulting in an average voxel size of approximately $\SI{0.1}{\meter} \times \SI{0.1}{\meter}$.

\noindent\textbf{Evaluation Metrics.}
We use the recall of the detector to study the  adversarial robustness of object detection (\cref{sec:adversarial robustness}). To accurately evaluate the attack performance, we also define Attack Success Rate (ASR) as the ratio of the number of hidden objects out of all targeted objects. 

\subsection{Adversarial Robustness of Object Detection}
\label{sec:adversarial robustness}
\begin{figure}[t]
    \centering

    \begin{subfigure}[t]{0.48\linewidth}
        \centering
        \includegraphics[width=.8\linewidth]{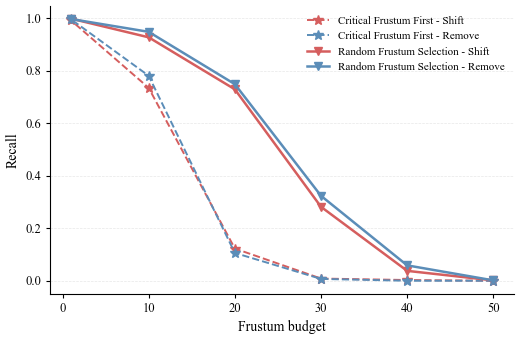}
        \caption{Robustness under attack strategies.}
        \label{fig:frustum_selection_perturbation_strategies}
    \end{subfigure}
    \hfill
    \begin{subfigure}[t]{0.48\linewidth}
        \centering
        \includegraphics[width=.8\linewidth]{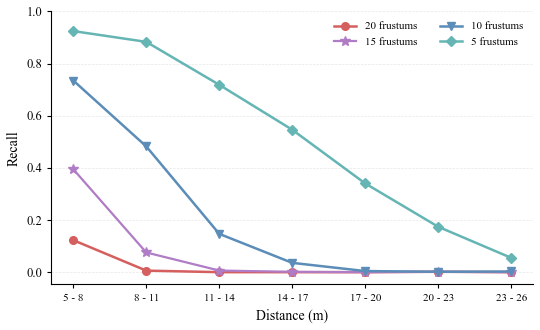}
        \caption{Robustness at different distances.}
        \label{fig:longer_distances}
    \end{subfigure}

    \vspace{1mm}

    \begin{subfigure}[t]{.9\linewidth}
        \includegraphics[width=\linewidth]{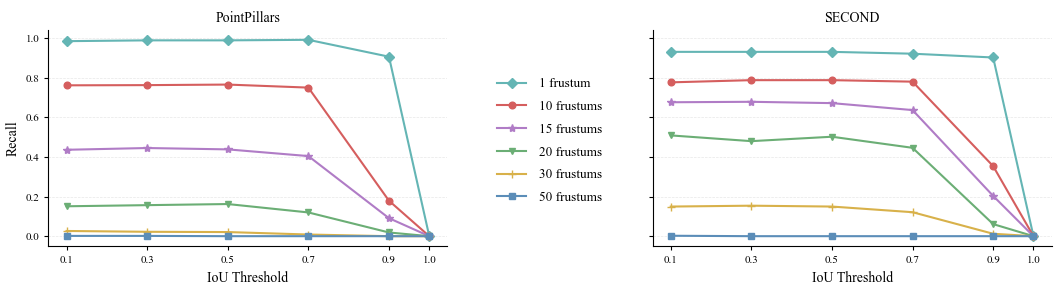}
        \caption{Detectors with different IoU thresholds.}
        \label{fig:iou_thres}
    \end{subfigure}

    \vspace{1mm}

    \begin{subfigure}[t]{\linewidth}
        \centering
        \includegraphics[width=\linewidth]{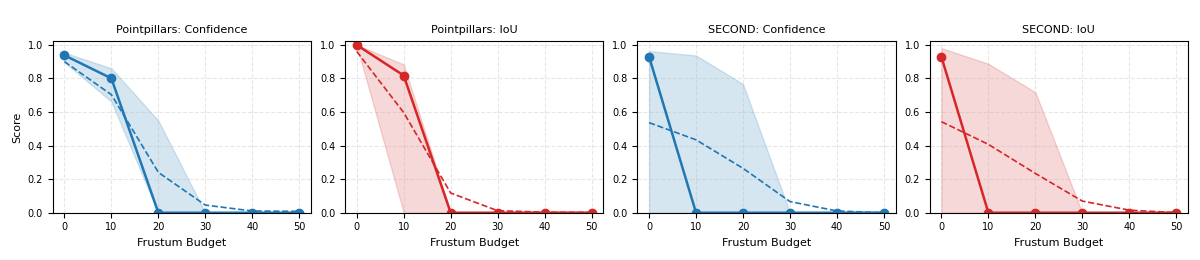}
        \caption{Robustness of detectors' internal parameters.}
        \label{fig:bbox_score_drop}
    \end{subfigure}

    \vspace{-2mm}
    \caption{Adversarial robustness of 3D object detection under the EFA attack.}
    \label{fig:frustum_level_analysis}
    \vspace{-5mm}
\end{figure}

\noindent\textbf{Robustness under Attack Strategies.}
\label{sec:ablation_selection}
We first evaluated the detection robustness under different \textit{EFA} attack selection and perturbation strategies. We defined four types of attacks  to test $Car$ objects predicted by PointPillars in KITTI dataset. As shown in \cref{fig:frustum_selection_perturbation_strategies}, recalls of all strategies decreased with increasing frustum budgets. The $Critical~Frustum~First$ selection strategies (guided by \textit{SALL}) decreased the recall to 0.0 using a frustum budget of 30, while $Random~Frustum~Selection$ required 50 frustums (the entire ROI) to achieve the same result. We find that the \textbf{selection strategy} is the dominant factor. Once the critical frustums are identified, the specific \textbf{perturbation strategy} ($\textit{Shift}$ or $\textit{Remove}$) has a negligible impact on the outcome. This is a key finding, as it simplifies the adversary's task to one of simply identifying and perturbing the most salient regions.

\noindent\textbf{Robustness at Different Distances.}
We tested the adversarial robustness of $Car$ objects detection from \SI{5}{m} to \SI{26}{m} using different frustum budgets as shown in \cref{fig:longer_distances}. It is noticeable that further objects are generally easier to attack. Using 5 frustums successfully hid \SI{94.5}{\percent} $Car$ objects at a distance of around \SI{25}{\meter}. This is because when ranges become longer and objects' 3D representation sparser, the detectors’ performance usually decreases significantly and therefore adversarially perturbed measurements have a more noticeable effect. 



\noindent\textbf{Robustness with IoU Thresholds.} 
We further studied the adversarial robustness of $Car$ object detection under our \textit{EFA} ($Critical~Frustum~First - Shift$) with different IoU thresholds. From \cref{fig:iou_thres}, we observe that when increasing the frustum budget, the recall falls. For $Cars$ at $IoU \ge 0.7$, we observe a signiﬁcant decline in recall for both detectors, with recall falling below 0.6 for most of the attacks showing that our \textit{EFA} attack is very effective in degrading the performance of object detection.

\noindent\textbf{Robustness of Internal Parameters.}
We further look into detector's internal parameters (confidence score and IoU values) under our EFA attacks. Both metrics quantify how EFA affects the detector’s internal prediction quality beyond the final attack outcome.
For both PointPillars and SECOND we report the median across attacked objects, with interquartile ranges shown as shaded regions in \cref{fig:bbox_score_drop}. Our evaluations show that increasing the frustum budget consistently drops both confidence and geometric overlap,  confirming EFA’s effectiveness from the detector’s internal prediction perspective.

\subsection{SALL-Guided Attack Effectiveness}
\begin{figure}[tb]
\centering
    \begin{subfigure}{0.32\linewidth}
        \includegraphics[width=\linewidth]{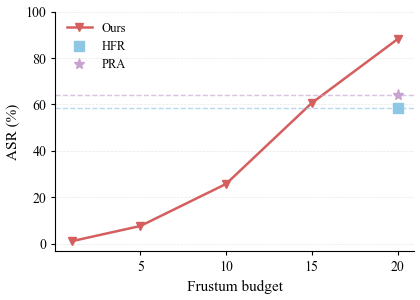}
        \caption{Attack KITTI dataset.}
        \label{fig:attack kitti car frustum}
    \end{subfigure}
    \begin{subfigure}{0.32\linewidth}
        \includegraphics[width=\linewidth]{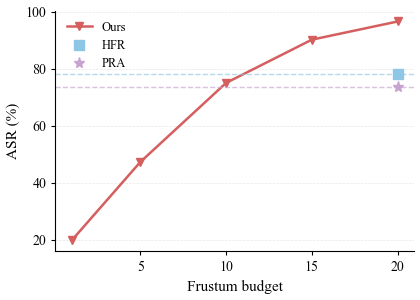}
        \caption{Attack nuScenes dataset.}
        \label{fig:attack nusc car frustum}
    \end{subfigure}
    \begin{subfigure}{0.32\linewidth}
        \includegraphics[width=\linewidth]{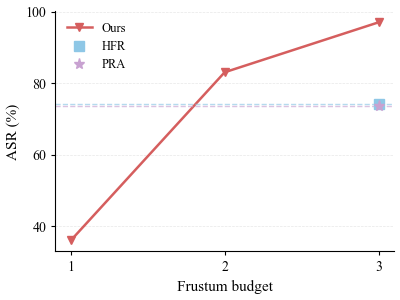}
        \caption{Attack small objects.}
        \label{fig:attack kitti pedestrian frustum}
    \end{subfigure}
    \vspace{-5pt}
    \caption{Saliency-guided attack performance on different datasets and object types. With the guidance of saliency maps, our attack consistently outperforms the baselines.}
    
    \label{fig:frustum less budgets}
    \vspace{-10pt}
\end{figure}

\noindent\textbf{Attacking Different Datasets.}
To validate our primary hypothesis, we deployed \textit{SALL}-guided attack ($Critical$ $Frustum~First$ selection with $Shift~Perturbation$) against $Car$ objects on both datasets. We compare our method against two baselines that mimic state-of-the-art (SOTA) brute-force frustum attacks: HFR~\cite{sato2024lidar} and PRA~\cite{cao2023you}. As shown in \cref{fig:attack kitti car frustum} and \cref{fig:attack nusc car frustum}, our ASR scales with the number of frustums perturbed. With the guidance of \textit{SALL}-generated saliency maps, our \textit{SALL}-guided attack consistently outperforms the baselines. Specifically, with a 20-frustum budget, our attack achieves a \SI{96.55}{\percent} ASR on nuScenes, an improvement of over 20 percentage points compared to HFR (\SI{78.29}{\percent}) and PRA (\SI{73.47}{\percent}). We observe a similar 30-point improvement on the KITTI dataset.

Crucially, our method achieves the same ASR as the baselines while requiring substantially fewer perturbed frustums: only 10 on nuScenes (a \textbf{\SI{50}{\percent} reduction}) and 
15 on KITTI (a \textbf{\SI{25}{\percent} reduction}). 
This shows that detector vulnerability is concentrated in a small subset of spatial regions, allowing the same attack success with substantially smaller perturbation budgets and potentially mitigating hardware overheating in physical LiDAR spoofing attacks~\cite{sato2025realism}.


\begin{wrapfigure}[10]{r}{0.19\textwidth}
\vspace{-20pt}
\centering
        \includegraphics[clip, trim=0.0cm 0cm 0.0cm 0cm, width=0.8\linewidth]{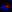}
    \vspace{-3pt}
    \caption{Universal saliency map of pedestrians.}
    \label{fig: kitti_pedestrian_map}

\end{wrapfigure}


\noindent\textbf{Attacking Small Objects.}
We evaluate our attack on small, difficult-to-detect objects, using $Pedestrian$ objects in KITTI as a test case. We generated a universal \textit{SALL} map for pedestrians (\cref{fig: kitti_pedestrian_map}) and applied our \textit{SALL}-guided attack. The results in \cref{fig:attack kitti pedestrian frustum} are stark. Our method achieves a \SI{97.00}{\percent} ASR with a tiny \textbf{3-frustum budget}. This is $\approx$ 23 percentage points higher than the HFR (\SI{74.25}{\percent}) and PRA (\SI{73.61}{\percent}) using the same budget. 
This indicates that brute-force frustum attacks~\cite{sato2024lidar, sato2025realism} are particularly inefficient on small objects, where only a few salient frustums are sufficient.

\begin{figure}[t]
    \centering

    \begin{subfigure}[t]{0.24\linewidth}
        \centering
        \includegraphics[width=\linewidth]{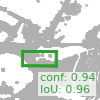}
        \caption{Car (benign).}
    \end{subfigure}
    \begin{subfigure}[t]{0.24\linewidth}
        \centering
        \includegraphics[width=\linewidth]{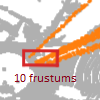}
        \caption{Car (attack).}
    \end{subfigure}
    \begin{subfigure}[t]{0.24\linewidth}
        \centering
        \includegraphics[width=\linewidth]{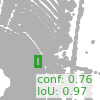}
        \caption{Pedestrian (benign).}
    \end{subfigure}
    \begin{subfigure}[t]{0.24\linewidth}
        \centering
        \includegraphics[width=\linewidth]{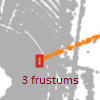}
        \caption{Pedestrian (attack).}
    \end{subfigure}

    \vspace{-5pt}
    \caption{EFA attack visualization. Orange points are perturbed, and gray points are unperturbed. Green box shows the target benign object, while red box marks the same region after attack, where the object is hidden.}
    \label{fig:efa_result_visualization}
    \vspace{-10pt}
\end{figure}

\noindent\textbf{Attack Visualization.}
\cref{fig:efa_result_visualization} provides qualitative  attack results on $Car$ and $Pedestrian$ objects, illustrating precise point manipulation and  resulting prediction removal. It is also clear that our method perturbs only a few critical frustums to attack $Car$ and $Pedestrian$ objects.

\subsection{Attack Transferability and Generalization}
\label{sec:ablation_model}
\noindent\textbf{Cross-Model Transferability.} To evaluate the transferability of our \textit{SALL} maps, we performed a cross-model attack on KITTI using a 30-frustum budget. We generated \textit{SALL} maps for PointPillars (\cref{fig:kitti_pp_car}) and SECOND (\cref{fig:kitti_second_car}) individually, as well as a joint map (\cref{fig:kitti_pp_second_car}). As shown in \cref{table:frustum_model_dependence}, the \textit{SALL} maps show strong transferability. Notably, a map generated from \textit{PointPillars} is effective at attacking \textit{SECOND}, improving ASR by $\approx$ 7 percentage points over the native \textit{SECOND} map (\SI{94.84}{\percent} vs. \SI{88.12}{\percent}). This suggests \textit{SALL} identifies shared object vulnerabilities rather than being dependent on model-specific artifacts. Furthermore, a joint map created by combining \textit{SALL} maps from both detectors improves performance on both models, achieving \SI{99.44}{\percent} on PointPillars. This confirms \textit{SALL}'s aggregation capability and suggests that maps become more universal as more detector architectures are included. Although runtime issues might arise due to multi-modal detectors which can be further avoided by running \textit{SALL} for each detector simultaneously and performing offline combination.

\begin{table*}[b]
\caption{Saliency map transferability across detectors(left) and datasets (right).}
\label{table:saliency_transfer}
\centering
\scriptsize
\setlength{\tabcolsep}{2pt}
\renewcommand{\arraystretch}{1.05}

\begin{subtable}[t]{0.475\textwidth}
\centering
\caption{Model dependence (ASR \%).}
\label{table:frustum_model_dependence}

\begin{tabular}{l S S}
\toprule
\multirow{2}{*}{\textbf{Map Src.}} & \multicolumn{2}{c}{\textbf{Target Model}} \\
\cmidrule(lr){2-3}
 & {\textbf{SECOND}} & {\textbf{PointPillars}} \\
\midrule
\textbf{SECOND} & 88.12 & 97.63 \\
\textbf{PointPillars} & 94.84 & 98.88 \\
\textbf{Joint} & 95.29 & 99.44 \\
\bottomrule
\end{tabular}
\end{subtable}
\hfill
\begin{subtable}[t]{0.475\textwidth}
\centering
\caption{Dataset dependence (ASR \%).}
\label{table:frustum_dataset_dependence}

\begin{tabular}{l S S}
\toprule
\multirow{2}{*}{\textbf{Map Src.}} & \multicolumn{2}{c}{\textbf{Target Data}} \\
\cmidrule(lr){2-3}
 & {\textbf{KITTI}} & {\textbf{nuScenes}} \\
\midrule
\textbf{KITTI} & 88.42 & 96.64 \\
\textbf{nuScenes} & 83.82 & 96.04 \\
\textbf{Joint} & 88.15 & 96.90 \\
\bottomrule
\end{tabular}
\end{subtable}

\end{table*}


\noindent\textbf{Cross-Dataset Transferability.} We performed a similar cross-dataset transferability test using a 20-frustum budget, attacking PointPillars on both KITTI and nuScenes. As shown in \cref{table:frustum_dataset_dependence}, the maps are highly transferable. A map learned from KITTI is highly effective on nuScenes, achieving a \SI{96.64}{\percent} ASR. Similar to our model-dependence findings, a joint saliency map aggregated from both KITTI and nuScenes further enhanced attack capability (achieving \SI{96.90}{\percent} ASR on nuScenes). This demonstrates \textit{SALL}'s robustness and its ability to generate highly representative saliency maps from diverse data sources.



\noindent\textbf{Extending SALL to Point-Level Attacks.} In addition to frustum-level attacks, we further explore \emph{point-level targeted attacks} based on saliency maps and compare against ORA \cite{hau2021object}. 
Following a trend similar to \cref{fig:frustum less budgets} , our \textit{SALL}-guided point attack consistently outperforms the baseline as the perturbation budget increases. With a 1200-point budget, it achieves \SI{91.07}{\percent} ASR on KITTI, improving over ORA by more than \SI{37}{\percent}, and shows a similar \SI{20}{\percent} ASR improvement on nuScenes. Moreover, it reaches the same ASR as the baseline with substantially fewer points: 600 on KITTI (\textbf{\SI{50}{\percent}} fewer) and 300 on nuScenes (\textbf{\SI{40}{\percent}} fewer). This shows that our \textit{SALL}-guided point-level attack significantly reduces the required spoofing signal, thereby lowering both attack complexity and potential hardware overheating issues.

\subsection{Runtime Overhead}
Universal saliency maps are generated once offline and reused during attack execution. Offline generation consists of \textit{Preprocessing} (3.8 ms), \textit{Adaptive Indexing} (14 ms), \textit{Aggregation} (3.5 ms), and \textit{IG Computation} (51.79 s), averaged over 10 scenes, on a 12 GB NVIDIA TITAN Xp GPU, with the latter dominating the one-time cost due to point density. Online execution requires only saliency lookup using compact maps (5.8 KB for $Cars$ and 0.3 KB for $Pedestrians$), introducing negligible overhead over existing LiDAR spoofing attacks. 

\subsection{Real-World Feasibility}
\begin{figure}[tb]
    \centering
    \begin{subfigure}[t]{0.48\linewidth}
        \centering
        \includegraphics[width=\linewidth]{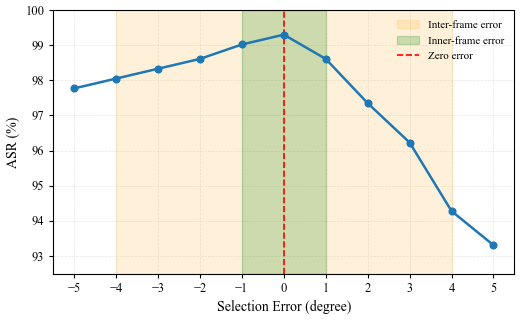}
        \caption{Attack with LiDAR spoofing errors.}
        \label{fig:selection_error_asr}
    \end{subfigure}
    \hfill
    \begin{subfigure}[t]{0.48\linewidth}
        \centering
        \includegraphics[width=0.8\linewidth]{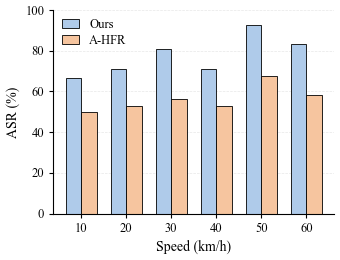}
        \caption{Speed-dependent perturbation constraints.}
        \label{fig:frustum_vehicle_speeds}
    \end{subfigure}
    \vspace{-2mm}
    \caption{Robustness of EFA under realistic physical-world constraints.}
    \label{fig:frustum_realworld_constraints}
    \vspace{-15pt}
\end{figure}

\noindent\textbf{Attack with LiDAR Spoofing Errors.}
SOTA synchronized LiDAR spoofing hardware is, by design, intended to precisely control the position of each spoofed point. This assumption has also been experimentally validated by HFR\cite{sato2024lidar}, which reported an inner-frame error of $\approx$ \SI{10}{cm} and an inter-frame error of $\approx$ \SI{35}{cm}. For front-near objects at \SIrange{5}{8}{m}, these correspond to $0.7^\circ\text{--}1.1^\circ$ and $2.5^\circ\text{--}4.0^\circ$ frustum selection error, respectively. We therefore evaluate EFA under selection errors from $-5^\circ$ to $5^\circ$. As shown in \cref{fig:selection_error_asr}, the ASR degrades only slightly: even within realistic inner-frame and inter-frame error ranges, the attack remains highly effective (over 98\% and 94\% ASR, respectively). This demonstrates EFA's strong robustness to realistic frustum selection errors. 

\noindent\textbf{Speed-Dependent Perturbation Constraints.}
We further evaluate our method under perturbation budgets derived from the 
point removal rates reported in A-HFR~\cite{sato2025realism}, which vary 
with vehicle speed. In A-HFR, higher driving speeds reduce the fraction of 
points that can be reliably manipulated. We simulate these speed-dependent 
constraints by applying the corresponding removal rates at speeds ranging 
from \SIrange{10}{60}{\kilo\meter\per\hour}.

We compare the original A-HFR strategy against a \textit{SALL}-guided 
variant using a 3-frustum budget for $Pedestrian$ objects. 
As shown in \cref{fig:frustum_vehicle_speeds}, incorporating 
\textit{SALL}-based critical-region selection consistently improves 
attack effectiveness across all speeds. The improvement is most pronounced at higher speeds 
(\SIrange{50}{60}{\kilo\meter\per\hour}), where perturbation budgets are 
more constrained. For example, at \SI{50}{\kilo\meter\per\hour}, 
\textit{SALL}-guided A-HFR achieves \SI{92.70}{\percent} success compared 
to \SI{67.38}{\percent} for the original method, an improvement of \SI{25}{\percent}. 

These results indicate that selecting structurally critical regions becomes 
increasingly important as perturbation budgets shrink, highlighting the 
efficiency of universal saliency priors under constrained conditions.

\section{Defense Analysis}



\noindent\textbf{Existing Defenses.}
Existing defenses such as \emph{CARLO} \cite{sun2020towards}, \emph{Shadow-Catcher} \cite{hau2021shadow} and \emph{3D-TC2} \cite{you2021temporal} primarily target injection-style anomalies, and their assumptions do not directly carry over to the object-hiding setting.
Object-hiding perturbations aim to suppress the 
detector’s output for a real object. This setting differs 
fundamentally from injection attacks in two respects: 
\emph{(i)} no bounding box may be produced for the target object, 
eliminating the anchor region required for frustum-based analysis; 
and \emph{(ii)} even under an oracle bounding box assumption, 
density-based checks become less discriminative when perturbations 
are sparse and localized, as in our saliency-guided setting. 
Adapting injection-focused defenses to this scenario would therefore 
require reasoning about missing objects or absence patterns, which 
goes beyond their original design assumptions.


\emph{Obstacle Detection} \cite{hau2022using} and \emph{FSD} \cite{cao2023you}  provided good insights of using 3D shadows for detecting object removal attacks, but they provided neither a quantitative analysis nor publicly available resources. We 
therefore implemented \emph{Obstacle Detection} to the best of our 
understanding and evaluated it on $7,481$ KITTI scenes. We find that even in benign scenarios, the method produces false alarms in $4,561$ scenes, yielding 
a false alarm rate of $60.96\%$. After manual inspection, we observed that naturally 
sparse background regions are frequently misinterpreted as shadow 
artifacts. This highlights the intrinsic difficulty of distinguishing 
structured removal from normal LiDAR sparsity, suggesting that 
current shadow-based heuristics face substantial challenges in this 
setting.


\noindent\textbf{Random Frustum Dropping.}
We further evaluate a  mitigation strategy based on random 
frustum dropping. After selecting target frustums using \textit{EFA}, we 
randomly discard a subset of frustums 
and measure the resulting ASRs. As shown in \cref{tab:random_frustum_dropping}, the 
\textit{SALL}-guided perturbation achieves an ASR of 
\SI{99.16}{\percent} without dropping. To reduce the ASR by 
approximately \SI{40}{\percent}, more than 10 out of 30 frustums must 
be removed. Such substantial random removal indiscriminately suppresses 
both informative and non-informative regions.

This behavior reflects a structural trade-off: modern 3D detectors 
concentrate discriminative evidence in a limited set of spatial regions. 
While this concentration improves detection efficiency, it also implies 
that perturbations targeting these regions are disproportionately 
effective. Conversely, defenses that indiscriminately discard large 
portions of the input risk degrading nominal detection performance 
unless the model is explicitly trained to distribute reliance more 
redundantly.

Overall, our findings suggest that input randomization 
alone is insufficient to mitigate saliency-aware perturbations. Model-level regularization or 
adversarial training could be potential alternatives for improving robustness.



\begin{table}[tb]
\caption{Random frustum dropping. To significantly reduce the ASR, more than 10 out of 30 frustums must be removed.}
\label{tab:random_frustum_dropping}
\centering
\footnotesize
\setlength{\tabcolsep}{4pt}
\renewcommand{\arraystretch}{1.05}
\begin{tabular}{lcccccc}
\toprule
\textbf{Dropped frustums} & 0 & 2 & 4 & 6 & 8 & 10 \\
\midrule
\textbf{ASR (\%)} & 99.16 & 98.19 & 94.00 & 85.63 & 72.38 & 59.14 \\
\bottomrule
\end{tabular}
\vspace{-2mm}
\end{table}

\section{Conclusion}
\label{sec:conclusion}

We propose \textbf{SALL}, an explainability-driven method for probing 
structural vulnerabilities in LiDAR-based 3D object detectors. 
By aggregating instance-level attributions into universal class-level 
saliency maps, we uncover stable spatial attribution patterns that 
persist across scenes, datasets, and detectors.
%
Unlike prior classification-based studies on isolated 3D objects, we evaluate full-scene autonomous driving perception under realistic sensing conditions.
Under geometry-consistent LiDAR perturbation constraints, targeting universally salient regions degrades detection performance with substantially smaller perturbation budgets than uniform strategies.
Across multiple detectors and datasets, saliency-guided perturbations achieve over \SI{95}{\percent} degradation rates, improving effectiveness by 
\SIrange{20}{30}{\percent} on large objects and approximately 
\SI{23}{\percent} on small objects compared to prior approaches 
\cite{cao2023you, sato2024lidar, sato2025realism}, while affecting 
\SIrange{25}{50}{\percent} fewer regions.

%
Our results show that 3D detectors concentrate discriminative evidence in spatially structured regions, creating a robustness–efficiency trade-off. More broadly, they demonstrate that attribution-driven adversarial analysis is a principled diagnostic tool for understanding and improving 3D detector robustness.

Future work could explore training strategies that encourage spatial 
redundancy or reduce saliency concentration, and extend attribution-driven 
analysis to more diverse benchmarks, additional sensors and multi-sensor fusion systems.

\bibliographystyle{splncs04}
\bibliography{11_references}
\newpage

\section{Ablation Study}
In realistic sensor spoofing scenarios, the SOTA LiDAR spoofer requires several hardware and software components to cooperate. This might lead to accuracy loss when applying our \textit{SALL}-generated maps. To evaluate the robustness and sensitivity of our SALL-guided attacks, we performed the following ablation studies. As a baseline, we set \textit{EFA} with 30 frustums to attack $Car$ objects predicted by PointPillars in KITTI dataset.

\subsection{Biased Object Extraction}
When the adversary uses \textit{EFA} to extract the target point cloud from the Region of Interest $R$, there may be extraction offsets, such as a biased extracted position and rotation of the target object.

\subsubsection{Heading Direction Offsets.}


\begin{table}[]
\caption{Influence of heading offsets of extracted target objects. Our attacks remain at high ASRs even with \SI{20}{\degree} heading offsets.}
\label{tab:heading_offset_influence}
\centering
\footnotesize
\setlength{\tabcolsep}{4.5pt}
\renewcommand{\arraystretch}{1.05}
\begin{tabular}{lccccc}
\toprule
\textbf{Rotation Degree} & \textbf{0} & \textbf{5} & \textbf{10} & \textbf{15} & \textbf{20} \\
\midrule
\textbf{Clockwise (+)}       & 99.16 & 99.16 & 98.19 & 96.93 & 95.12 \\
\textbf{Anti-clockwise (-)}  & 99.16 & 98.61 & 97.49 & 96.79 & 94.14 \\
\bottomrule
\end{tabular}
\vspace{-2mm}
\end{table}
We first measure the influence of the heading direction offsets of the target object. The results in \cref{tab:heading_offset_influence} show that even with clockwise and anticlockwise heading direction offsets, our \textit{EFA} still manages to maintain a high attack performance. Specifically, \textit{EFA} can tolerate a 20-degree object heading offset with an over \SI{94}{\percent} ASR. This result is significant: it demonstrates the heading direction robustness of our \textit{SALL}-guided attacks.

\subsubsection{Object Position Offsets.}
To explore the influence of the target object's position offsets, we evaluated an offset distance of \SI{20}{\cm} on each axis of \textit{X}, \textit{Y}, and \textit{Z}. As shown in \cref{tab:position_offsets}, our \textit{SALL}-guided attack is robust to both positive and negative interference on \textit{X}, \textit{Y}, \textit{Z} axes. Comparing with the no-offset case (\SI{99.16}{\percent}), there is only a maximum of \SI{2}{\percent} ASR reduction on X and Y axes. Notably, there is no performance influence when the target object is offset on the Z axis which might have been mitigated by our \textbf{2D Indexing} approach (\cref{sec:EFA}).



\begin{table}[]
\caption{Influence of object position offsets. Our attacks are robust to object position offsets on different axes.}
\label{tab:position_offsets}
\centering
\footnotesize
\setlength{\tabcolsep}{5pt}
\renewcommand{\arraystretch}{1.05}
\begin{tabular}{lccc}
\toprule
\textbf{Position} & \textbf{No offset} & \textbf{Positive (+)} & \textbf{Negative (-)} \\
\midrule
\textbf{X} & 99.16 & 99.86 & 97.49 \\
\textbf{Y} & 99.16 & 98.74 & 97.21 \\
\textbf{Z} & 99.16 & 99.16 & 99.16 \\
\bottomrule
\end{tabular}
\vspace{-2mm}
\end{table}

\subsection{2D Indexing Influence}
To explore the sensitivity of our \textit{EFA} in terms of \textbf{2D Indexing}, we added a 2-voxel offset ($\approx$ 20cm) on each of four directions: forward, backward, left and right. As shown in \cref{tab:2d_indexing_offsets}, most offsets don't have much influence on \textit{EFA}'s attack performance. Noteably, offsets in left and right directions can lead to larger attack performance reduction than forward and backward directions which is worth further exploration and development of mitigation strategies.



\begin{table}[t]
\caption{Influence of 2D indexing offsets. Most offsets in different directions have little influence on attack performance.}
\label{tab:2d_indexing_offsets}
\centering
\footnotesize
\setlength{\tabcolsep}{6pt}
\renewcommand{\arraystretch}{1.05}
\begin{tabular}{lc}
\toprule
\textbf{Offset Direction} & \textbf{ASR (\%)} \\
\midrule
\textbf{No Offsets} & 99.16 \\
\textbf{Left}       & 92.89 \\
\textbf{Right}      & 72.25 \\
\textbf{Forward}    & 98.33 \\
\textbf{Backward}   & 98.05 \\
\bottomrule
\end{tabular}
\vspace{-2mm}
\end{table}

\subsection{Positive vs. Negative Contributions}
In our saliency maps, besides positive contributions, there are also some negative contributions (blue pixels in \cref{fig:saliency map visualization}). Targeting $Car$ objects in KITTI dataset, we use positive and negative contributions to guide \textit{EFA} and compare the attack performance using different frustum budgets from 1 to 50. 
As shown in \cref{fig: positive_negative_contributions}, negative contributions are also effective but not as good as positive contributions at hiding objects. When using more than 40 frustums, negative contributions share the same attack performance as positive contributions.

\begin{minipage}{0.45\textwidth}
\includegraphics[width=\linewidth]{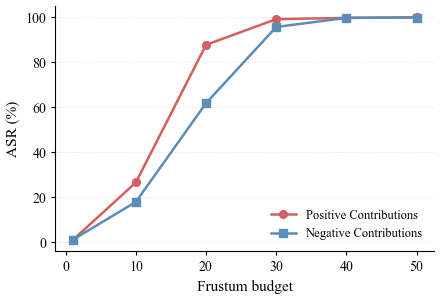}
    \captionof{figure}{Effectiveness of contributions. Positive contributions are more effective than Negative contributions.}
    \label{fig: positive_negative_contributions}
\end{minipage}
\hfill
\begin{minipage}{0.45\textwidth}    \includegraphics[width=\linewidth]{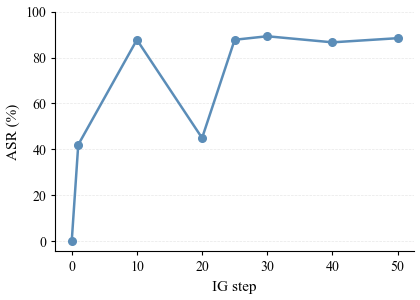}
    \captionof{figure}{Influence of IG steps. The attack performance becomes stable after step 25.}
    \label{fig: IG_steps}
\end{minipage}

\subsection{IG Steps}
IG step dominates the overall runtime of the saliency map generation. More IG steps require more runtime in \textit{Integrated Gradient Computation}. Using different IG steps from 0 to 50, we generated saliency maps for $Car$ objects in KITTI dataset. Then saliency maps at different IG steps are used to guide \textit{EFA} and calculate ASRs with the same frustum budget of 20 in \cref{fig: IG_steps}. Overall, increasing IG steps doesn't necessarily increase the effectiveness of saliency maps especially when IG steps are higher than 25. Despite instability, using 10 IG steps can also achieve a relatively higher attack performance which indicates a potential direction for reducing the runtime of saliency map generation.


\subsection{Base Points}
\label{sec:basemaps}

\noindent We studied the effectiveness of base points while generating saliency maps. We defined two types of base points to generate saliency maps: $Zero$ (which is the LiDAR origin) and $Nearest~Corner$ of the target bounding box. For each saliency map, ASR is calculated while performing \textit{EFA} attacks on front-near $Car$ objects with different frustum budgets. As shown in \cref{tab:basepoints}, saliency maps generated based on $Zero$ seem similarly effective as $Nearest~Corner$. This proves that saliency map generation doesn't rely on the physics of LiDAR rays but learn from consistent perturbations.

\begin{table}[]
\caption{Influence of base points. Saliency maps generated based on \textit{Zero} remain similarly effective to \textit{Nearest Corner} when attacking frustums.}
\label{tab:basepoints}
\centering
\footnotesize
\setlength{\tabcolsep}{4.5pt}
\renewcommand{\arraystretch}{1.05}
\begin{tabular}{lcccccc}
\toprule
\textbf{Frustum Budget} & \textbf{50} & \textbf{40} & \textbf{30} & \textbf{20} & \textbf{10} & \textbf{1} \\
\midrule
\textbf{Zero}           & 100.00 & 100.00 & 98.88 & 90.66 & 29.43 & 1.39 \\
\textbf{Nearest Corner} & 100.00 & 99.72  & 99.16 & 87.73 & 26.50 & 0.84 \\
\bottomrule
\end{tabular}
\vspace{-2mm}
\end{table}
\end{document}